\documentclass{article}

\usepackage{microtype}
\usepackage{graphicx}
\usepackage{subfigure}
\usepackage{booktabs} 
\usepackage{makecell}
\usepackage{colortbl}  
\usepackage{threeparttable}
\usepackage{hyperref}
\usepackage{multirow}

\newcommand{\fullname}{Autonomy-of-Experts}
\newcommand{\name}{AoE}
\newcommand{\argtopk}{\texttt{argtopK}}
\newcommand{\plusequ}{\mathrel{+}=}
\usepackage{tikz}  
\newcommand{\circlednum}[1]{%
  \tikz[baseline, anchor=base]{\node[draw,circle,inner sep=0.2pt, font=\fontsize{9}{10}\selectfont] {#1};}%
}

\newcommand{\circledNum}[1]{%
  \tikz[baseline, anchor=base]{\node[draw,circle,inner sep=0.1pt, font=\fontsize{7.5}{10}\selectfont] {#1};}%
}
\usepackage[accepted]{icml2025}

\usepackage{amsmath}
\usepackage{amssymb}
\usepackage{mathtools}
\usepackage{amsthm}
\usepackage{bbm}

\usepackage[capitalize,noabbrev]{cleveref}

\theoremstyle{plain}

\theoremstyle{definition}

\theoremstyle{remark}

\usepackage[textsize=tiny]{todonotes}

\icmltitlerunning{Autonomy-of-Experts Models}

\begin{document}

\twocolumn[
\icmltitle{Autonomy-of-Experts Models}




\begin{icmlauthorlist}
\icmlauthor{Ang Lv}{ruc}
\icmlauthor{Ruobing Xie}{tencent}
\icmlauthor{Yining Qian}{seu}
\icmlauthor{Songhao Wu}{ruc}
\icmlauthor{Xingwu Sun}{tencent,um}
\icmlauthor{Zhanhui Kang}{tencent}
\icmlauthor{Di Wang}{tencent}
\icmlauthor{Rui Yan}{ruc,moe,whu}
\end{icmlauthorlist}

\icmlaffiliation{ruc}{Gaoling School of Artificial Intelligence, Renmin University of China}
\icmlaffiliation{tencent}{Large Language Model Department, Tencent}
\icmlaffiliation{seu}{Southeast University, China}
\icmlaffiliation{whu}{School of Artificial Intelligence, Wuhan University}
\icmlaffiliation{moe}{Engineering Research Center of Next-Generation Intelligent Search and Recommendation, MoE}
\icmlaffiliation{um}{University of Macau}
\icmlcorrespondingauthor{Ruobing Xie}{xrbsnowing@163.com}
\icmlcorrespondingauthor{Rui Yan}{ruiyan@ruc.edu.cn}

\icmlkeywords{Autonomy-of-Experts models, Large language models}

\vskip 0.3in
]



\printAffiliationsAndNotice{}  

\begin{abstract}
Mixture-of-Experts (MoE) models mostly use a router to assign tokens to specific expert modules, activating only partial parameters and often outperforming dense models. 
We argue that the separation between the router's decision-making and the experts' execution is a critical yet overlooked issue, leading to suboptimal expert selection and ineffective learning. 
To address this, we propose \fullname\ (\name), a novel MoE paradigm in which experts autonomously select themselves to process inputs.
\name\ is based on the insight that an expert is aware of its own capacity to effectively process a token, an awareness reflected in the scale of its internal activations.
In \name, routers are removed; instead, experts pre-compute internal activations for inputs and are ranked based on their activation norms.
Only the top-ranking experts proceed with the forward pass, while the others abort. 
The overhead of pre-computing activations is reduced through a low-rank weight factorization.
This self-evaluating-then-partner-comparing approach ensures improved expert selection and effective learning.
We pre-train language models having 700M up to 4B parameters, demonstrating that \name\ outperforms traditional MoE models with comparable efficiency.
The code is available at \url{https://github.com/trestad/Autonomy-of-Experts}
\end{abstract}

\begin{figure}[!t]
\begin{center}
\centerline{\includegraphics[width=\columnwidth]{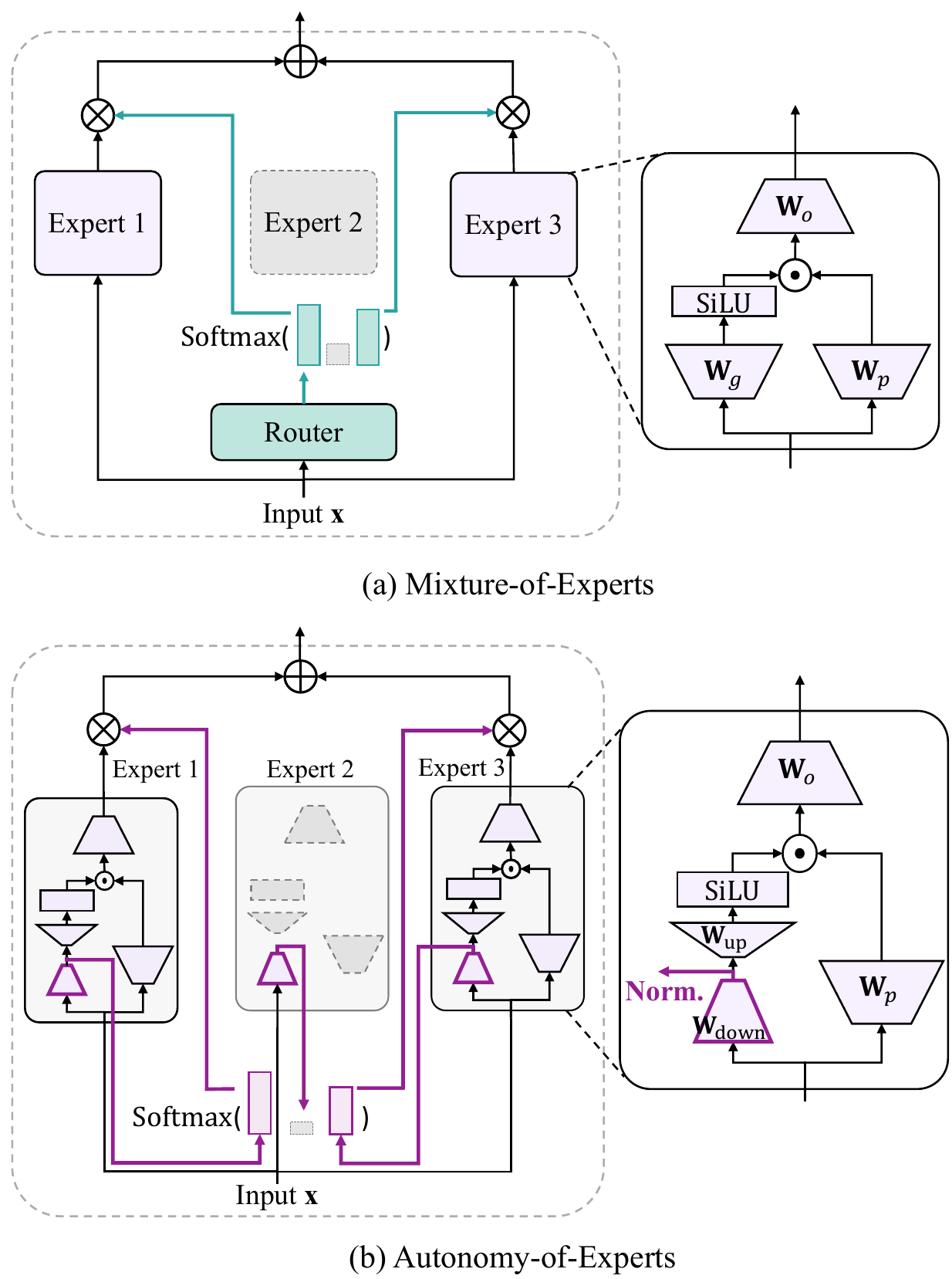}}
\caption{Comparison between traditional MoE and \name.
    Arrows indicate data flow, while shadowed modules represent unused parameters or variables.
    (a) Traditional MoE models use a router to assign tokens to specific experts. 
    This separation between the router‘s decision-making and the experts' execution leads to suboptimal expert selection and ineffective learning.
    (b) In an \name\ model, experts operate autonomously. 
    They are ranked based on their internal activation norms, and only the top-activated experts continue processing, while the others are terminated. 
    The \name\ expert architecture is modified to maintain efficiency.
    }
\label{fig:teaser}
\end{center}
\end{figure}

\section{Introduction}

Large language models (LLM) built on Mixture-of-Experts techniques (MoE,~\citealp{shazeer2017,lepikhin2021gshard,fedus2022switchtransformersscalingtrillion}) have gained increasing research and industrial attention~\cite{jiang2024mixtralexperts,dai2024deepseekmoeultimateexpertspecialization,qwen_moe,sun2024hunyuanlargeopensourcemoemodel}.
The core idea of MoE in LLMs involves dividing a large feed-forward network (FFN) into smaller FFNs, known as experts, and activating different experts' parameters for different inputs.
The decision on which experts process which inputs are made by a router, typically an MLP-based classifier.
Compared to dense models, MoE models are more efficient due to their sparse activation, and their ability to flexibly combine expert knowledge enhances downstream performance.

A critical issue in MoE is often overlooked: the separation between the router's decision-making and the experts' execution. 
The router cannot directly assess the experts' abilities, making its selection of an expert essentially a prediction without available labels.
If the router makes an incorrect prediction, the chosen expert may struggle to process the tokens effectively, leading to increased training loss. 
To reduce the loss, the expert might adapt its parameters to handle these tokens, potentially conflicting with its original expertise. 
Alternatively, the router must learn to make better decisions through trial and error, as it lacks awareness of which experts are best suited for specific tasks, thereby wasting many training steps.

To address these challenges, we propose a novel MoE paradigm—\fullname\ (\name). 
\name\ allows experts to decide whether to process inputs themselves. 
This design is based on the observation that experts are aware of their ability to handle inputs, an awareness reflected in the scale of their internal activations. 
Building on this insight, we enable all experts in an \name\ layer to process every token and cache their internal activations. 
For each token, experts are ranked by their internal activation norms, with only the top-ranked experts continuing to process the token using the cache, while the others terminate the process. 
The additional overhead from caching and computations of unused experts is mitigated by factorizing the experts' weights, which compresses the inputs into low-dimensional vectors for efficient caching.
Due to the autonomy of the experts, the router is eliminated.
Figure~\ref{fig:teaser} presents a comparative overview of traditional MoE and \name\ models.

We pre-train \name\ language models with up to 4 billion parameters, and they outperform traditional MoE models on downstream tasks with comparable efficiency.
We provide a comprehensive analysis of \name\ to highlight its advantages.
These advantages include improved expert selection, more specialized experts, and more effective training, all of which contribute to better downstream performance.

\section{Background: Mixture-of-Experts (MoE)}

We focus on sparse MoE models, treating each feed-forward network (FFN) module as an expert.
Each FFN, or expert, is expected to possess diverse and distinct abilities, enabling the model to process inputs effectively by activating only the experts with the necessary capabilities, thereby improving efficiency.
Some studies~\cite{chen-etal-2024-fortify,lin2024mixture} on dense MoE do not reduce the parameter activation ratio, which is not the primary concern of this paper.
In this paper, when we refer to MoE, we mean sparse MoE.

\begin{algorithm}[tb]
   \caption{A working pipeline of an MoE layer}
   \label{alg:moe}
\begin{algorithmic}[1]
   \INPUT A hidden state $\mathbf{x}\in\mathbb{R}^{d_{\text{model}}}$, number of experts $n$. 
   \OUTPUT The layer output $\mathbf{h} \in \mathbb{R}^{d_{\text{model}}}$, initialized as zeros.
   \STATE $\mathbf{p} = R(\mathbf{x})$
   \hfill \text{// $\mathbf{p} \in\mathbb{R}^{n}$}
   \STATE $\mathbf{I} = \argtopk(\mathbf{p})$  \hfill \text{// $\mathbf{I} \in\mathbb{R}^{K}$}
   \STATE $\mathbf{\hat{p}} = \texttt{Softmax}(\mathbf{p}[\mathbf{I}])$ 
   \hfill \text{// $\mathbf{\hat{p}} \in\mathbb{R}^{K}$}
   \FOR{$i=1$ {\bfseries to} $n$}
   \IF{$i \in \mathbf{I}$}
   \STATE $\mathbf{h} \plusequ \mathbf{\hat{p}}[i] \cdot E_{i}(\mathbf{x})$
   \ENDIF
   \ENDFOR
\end{algorithmic}
\end{algorithm}

\begin{table*}[t]
    \caption{
We remove routers from pre-trained MoE-LLMs and select experts during inference based on the internal activation norms of specific nodes in the computational graph. 
The accuracy on two challenging tasks is reported, along with the time cost (in minutes) for 8×A800-80G GPUs, which is given in parentheses. 
Without parameter updates, we can largely preserve accuracy under certain nodes, but this rudimentary approach requires significant improvements in efficiency.
    }
    \label{tab:pre}
    \vskip 0.1in
\begin{center}
\begin{small}
    \begin{tabular}{c|c|c|c|c}
    \toprule
    \multirow{2}{*}{\shortstack{Node for Norm\\ Calculation}} & \multicolumn{2}{c|}{MMLU (5-shot)} & \multicolumn{2}{c}{ARC-C (5-shot)} \\\cmidrule(lr){2-5}
    & Mixtral $8\times7$B  & Phi-3.5-MoE-ins. & Mixtral $8\times7$B & Phi-3.5-MoE-ins. \\\midrule
    $\mathbf{x}\mathbf{W}_{g}$  & 64.23 (42.70) & 29.43 (33.05) & 50.43 (4.40)  & 28.84 (3.47) \\
    $\mathbf{x}\mathbf{W}_{p}$ & 62.06 (42.73) & 34.60 (33.05) & 53.41 (4.40) & 40.36 (3.47) \\
    $\texttt{SiLU}(\mathbf{x}\mathbf{W}_{g})$ & 61.71 (43.88) & 38.03 (34.32) & 58.79 (4.51)  & 47.53 (3.60) \\
    $\texttt{SiLU}(\mathbf{x}\mathbf{W}_{g}) \odot \mathbf{x}\mathbf{W}_{p}$ & 66.64 (75.53) & 27.89 (52.60) & 58.79 (6.27) & 35.32 (5.42) \\
    Experts' Final Outputs & 66.66 (76.15) & 29.69 (69.20) & 58.62 (7.42) & 36.35 (7.07) \\
    \midrule
    \midrule
    Performance w. Router & 70.35 (24.30) & 78.20 (14.53) & 62.12 (2.50) & 67.41 (1.60) \\\bottomrule
    \end{tabular}
    \end{small}
    \end{center}
\end{table*}

MoE-based LLMs~\cite{jiang2024mixtralexperts,dai2024deepseekmoeultimateexpertspecialization,qwen_moe,jamba,sun2024hunyuanlargeopensourcemoemodel,abdin2024phi3technicalreporthighly} typically follow the FFN design in the Llama models~\cite{touvron2023llamaopenefficientfoundation} as an expert module.
The $i$-th expert within a specific layer can be formulated as: 
\begin{equation}
   E_{i}(\mathbf{x}) = \left(\texttt{SiLU}(\mathbf{x}\mathbf{W}^{i}_{g}) \odot (\mathbf{x} \mathbf{W}^{i}_{p})\right)  \mathbf{W}^{i}_{o} ,
\end{equation}
where $\mathbf{x} \in \mathbb{R}^{d_{\text{model}}}$ is the input hidden state; $\mathbf{W}^{i}_{g}, \mathbf{W}^{i}_{p} \in \mathbb{R}^{d_{\text{model}} \times d_{\text{ffn}}}$, and $\mathbf{W}^{i}_{o} \in \mathbb{R}^{d_{\text{ffn}} \times d_{\text{model}}}$ are the expert weights.
This paper focuses on this classical FFN formulation.

A router (or gate) $R$ determines which expert processes which hidden state. 
Many studies have proposed various routing strategies, such as token choosing top experts~\cite{shazeer2017,lepikhin2021gshard}, expert choosing top tokens~\cite{zhou2022mixtureofexperts,pmlr-v202-zhou23c}, dynamic expert calls~\cite{mod,gong-etal-2024-mixture}, and refining expert selection by solving mathematical problems~\cite{pmlr-v139-lewis21a,pmlr-v162-clark22a}, among others.
Without loss of generality, our discussion focuses on token choosing the Top-$K$ experts~\cite{shazeer2017,lepikhin2021gshard}, but our experiments consider various strategies.
Algorithm~\ref{alg:moe} presents a working pipeline of an MoE layer with a total of \( n \) experts.
The ``$[i]$'' notation in the algorithm follows Python syntax, indicating the selection of the $i$-th element in a vector or a matrix.

A challenge faced by MoE is the imbalanced expert load. 
MoE routers tend to disproportionately favor specific experts, resulting in suboptimal parameter utilization. 
\citet{fedus2022switchtransformersscalingtrillion} incorporate a load-balancing loss, controlled by a hyperparameter weight, $\alpha_{\text{aux}}$, to ensure that each expert receives a similar load for a batch $\mathcal{B}$ with $T$ tokens:
\begin{equation}
    \begin{aligned}
        &\mathcal{L}_{\text{aux}} = \alpha_{\text{aux}}\cdot n \cdot \sum^{n}_{i=1} \mathbf{f}_{i} \cdot \mathbf{P}_{i}, \text{\normalsize{where}} \\
        &\mathbf{f}_{i} = \frac{1}{T}\sum_{\mathbf{x}\in\mathcal{B}} \mathbbm{1}\left\{i \in \argtopk\left(R\left(\mathbf{x}\right)\right)\right\},\\
        &\mathbf{P}_{i} = \frac{1}{T}\sum_{\mathbf{x}\in\mathcal{B}} \texttt{Softmax}\left(R\left(\mathbf{x}\right)\right)[i].
    \end{aligned}
    \label{eq:aux}
\end{equation}
Several variants of this auxiliary loss are proposed~\cite{zuo2022taming,wang2024hmoeheterogeneousmixtureexperts,wang2024auxiliarylossfreeloadbalancingstrategy,huang2024hardertasksneedexperts}, sharing the same load-balancing goal.
Therefore, our discussion focuses on the balancing loss presented above.

Several studies~\cite{roller2021hash,gururangan-etal-2022-demix,ren2023pangusigmatrillionparameterlanguage,fan2020englishcentricmultilingualmachinetranslation} classify tokens based on prior knowledge—such as domain, language, or hash mapping—and assign them to fixed experts. 
While they do not use explicit routers, they differ significantly from \name\ in many respects.
Most importantly, their expert selection is not determined by the experts themselves, leaving the separation between decision-making and execution unaddressed.
\citet{pham2024competesmoe} use the norm of expert final outputs as the label for router logits. This method shares the concept with ours, where the activation norm represents expertise; however, it incurs dense activation across all experts and does not address the separation issue we highlighted.

\section{Method}

We begin by introducing preliminary experiments that motivate the development of \fullname\ (\name) in Section~\ref{sec:ob}. 
In Section~\ref{sec:method}, we refine the straightforward implementation from the preliminary experiments, improving the expert architecture to address efficiency concerns and, finally, deriving the \name\ method.

\subsection{An Insight: Experts ``Know'' What They Know}
\label{sec:ob}

We present the experiment that motivated the development of \name\ models.

\citet{geva-etal-2021-transformer} interpret FFN layers as key-value memory networks, where inputs are projected into a ``key'' vector (e.g., $\left(\texttt{SiLU}(\mathbf{x}\mathbf{W}{g}) \odot (\mathbf{x} \mathbf{W}{p})\right)$). 
The ``key'' vector retrieves knowledge or abilities stored in the parameters through a key-value matching mechanism (e.g., multiplying by $\mathbf{W}_{o}$). 
If the experts can effectively handle the input, the ``key'' should be highly activated, allowing for effective retrieval. 
Note that this example is purely analogical; there are no defined rules to determine which internal activations behave more like the ``key'' and which behave more like the ``value,'' as models are not trained with constraints that would regularize these roles.

Inspired by \cite{geva-etal-2021-transformer}, we conducted preliminary experiments to explore whether experts in pre-trained MoE-LLMs ``know'' their capabilities—that is, whether the scale of their activation norms reflects their ability to handle specific inputs. 
Specifically, for a given pre-trained MoE-LLM, we remove all routers and let \textit{every} expert within a layer to process each input up to a specific ``pause'' node in the computational graph (e.g., after $\mathbf{x}$ is multiplied by $\mathbf{W}_{g}$). 
We then ranked the experts based on the $L^{2}$ norm of their activations at the node.\footnote{We also evaluated the $L^1$ and $L^\infty$ norms, but these performed worse than the $L^2$ norm, as detailed in Appendix~\ref{apx:norms}.}
The top-$K$ experts continue the forward pass from the pause node to generate the final MoE outputs, while the others are terminated. 
We conducted 5-shot tests on Mixtral $8\times7$B~\cite{jiang2024mixtralexperts} and Phi-3.5-MoE-instruct~\cite{abdin2024phi3technicalreporthighly} using MMLU~\cite{hendrycks2021measuringmassivemultitasklanguage} and ARC-Challenge~\cite{clark2018thinksolvedquestionanswering}, and investigated how much of the performance of these LLMs can be preserved using this expert-selection strategy.

Regarding which node to use for calculating the activation norm, we conducted several trials.
The accuracy scores under various setups are shown in Table~\ref{tab:pre}. 
We also report the time taken on 8$\times$A800-80G.
The test code is based on the LM Evaluation Harness~\cite{eval-harness} with a batch size of 50.
Experiments across different models and tasks reveal that the optimal nodes for preserving the performance of a \textit{pre-trained} LLM vary. 
This finding supports the earlier assertion that, in a \textit{pre-trained} LLM, there is no predetermined node whose norm best reflects experts' underlying abilities.
Notably, this experiment does not update any parameters and is conducted under out-of-distribution inference behavior, i.e., without routers.
Despite this, performance preservation reaches up to 95\% for Mixtral and 71\% for Phi-3.5.

These preliminary results motivate us to train an MoE model \textit{from scratch} with an explicit designation of the node for expert selection. 
We expect that the model will naturally learn to represent its awareness of its capabilities through the norm of the designated node. 
Such an approach could effectively address the separation between the router's decision-making and the experts' execution—a challenge inherent in traditional MoE models.

\begin{algorithm}[tb]
   \caption{A working pipeline of an \name\ layer}
   \label{alg:aoe}
\begin{algorithmic}[1]
   \INPUT A hidden state $\mathbf{x}\in\mathbb{R}^{d_{\text{model}}}$, number of experts $n$. Initialize the activation cache $\mathbf{C} \in \mathbf{R}^{n \times d_{\text{low}}}$ and $\mathbf{p} \in \mathbb{R}^{n}$ as all zeros.
   \OUTPUT The layer output $\mathbf{h} \in \mathbb{R}^{d_{\text{model}}}$, initialized as zeros.
   \STATE \text{// In practice, we replace the following loop with a}
   \STATE \text{// single matrix multiplication (see Eq.~\ref{eq:fast}) for efficiency.}
   \FOR{$i=1$ {\bfseries to} $n$}   
   \STATE $\mathbf{C}[i] = \mathbf{x} \mathbf{W}^{i}_{\text{down}} $ 
   \hfill \text{// $\mathbf{C}[i] \in\mathbb{R}^{d_{\text{low}}}$}
   \ENDFOR
   \STATE $\mathbf{p} = \texttt{L2-Norm}(\mathbf{C},\ \text{dim=-1})$
   \hfill \text{// $\mathbf{p} \in\mathbb{R}^{n}$}
   \STATE $\mathbf{I} = \argtopk(\mathbf{p})$
   \hfill \text{// $\mathbf{I} \in\mathbb{R}^{K}$}
   \STATE $\mathbf{\hat{p}} = \texttt{Softmax}(\mathbf{p}[\mathbf{I}])$
   \hfill \text{// $\mathbf{\hat{p}} \in\mathbb{R}^{K}$}
   \FOR{$i=1$ {\bfseries to} $n$}
   \IF{$i \in \mathbf{I}$}
   \STATE $\mathbf{h} \plusequ \mathbf{\hat{p}}[i] \cdot
  \left((\texttt{SiLU}(\mathbf{C}_{i} \mathbf{W}^{i}_{\text{up}} ) \odot (  \mathbf{x}\mathbf{W}^{i}_{p})) \mathbf{W}^{i}_{o}\right)$
   \ENDIF
   \ENDFOR
\end{algorithmic}
\end{algorithm}

\subsection{\fullname\ (\name)}
\label{sec:method}

The following paper centers on using the norm of $\mathbf{xW}_{g}$ to guide expert selection in our new MoE language models pre-trained from scratch.
There is no technical difference or challenge in applying our method to any other node, regardless of the architecture.
However, utilizing nodes other than $\mathbf{xW}_{g}$ or $\mathbf{xW}_{p}$ is not cost-effective.

The efficiency of the rudimentary method in Section~\ref{sec:ob} must be improved.
The primary overhead arises from all experts computing activations for a given token, even though not all results contribute to the final MoE output.
Additionally, large $d_{\text{ffn}}$-dimensional activations (14,336 for Mixtral $8\times 7$B and 6,400 for Phi-3.5-MoE) at the pause node are cached, leading to significant memory usage.

A factorization of the $\mathbf{W}_{g}$ matrix can address these two issues. 
We decompose $\mathbf{W}_{g}$ into two low-rank matrices: $\mathbf{W}_{\text{down}} \in \mathbb{R}^{d_{\text{model}} \times d_{\text{low}}}$ and $\mathbf{W}_{\text{up}} \in \mathbb{R}^{d_{\text{low}} \times d_{\text{wide}}}$, where $d_{\text{low}} < d_{\text{model}} < d_{\text{wide}}$.
The $i$-th \name\ expert can be formulated as:
\begin{equation}
E_{i}(\mathbf{x}) = \left(\texttt{SiLU}\left(\mathbf{x}\mathbf{W}^{i}_{\text{down}}\mathbf{W}^{i}_{\text{up}} \right) \odot \left(\mathbf{x}\mathbf{W}^{i}_{p}\right)\right)\mathbf{W}^{i}_{o},
\end{equation}
where $\mathbf{W}^{i}_{p} \in \mathbb{R}^{d_{\text{model}} \times d_{\text{wide}}}$, and $\mathbf{W}^{i}_{o} \in \mathbb{R}^{d_{\text{wide}} \times d_{\text{model}}}$.

Algorithm~\ref{alg:aoe} formulates the pipeline within an \name\ layer.
In each expert, $\mathbf{W}_{\text{down}}$ first compresses the input vectors into low-dimensional activations. 
These activations are cached as $\mathbf{C}$, and their $L^{2}$ norms are used to rank the experts. 
Given an input, the experts with the top-$K$ norms use the cache to continue the forward computation within the expert, while unchosen experts abort processing. 
The compressed activations significantly reduce both the cache size and the computational overhead from unselected experts. 
This factorization does not impair the model's expressiveness, as the weights are inherently low-rank in large language models~\cite{li2018measuring,aghajanyan-etal-2021-intrinsic,hu2022lora}.

Furthermore, to enhance efficiency, the loop for calculating the activation cache (Line 2 in Algorithm~\ref{alg:aoe}) can be eliminated by combining the $\mathbf{W}^{i}_{\text{down}}$ matrices of all experts into a single large matrix. 
This allows the cache to be obtained through a single multiplication:
\begin{equation}
    \begin{aligned}
    \mathbf{\hat{W}}_{\text{down}} &= [\mathbf{W}^{1}_{\text{down}}, \cdots, \mathbf{W}^{n}_{\text{down}}] \in \mathbb{R}^{d_{\text{model}}\times (n d_{\text{low}})}\\
        \mathbf{C} &= \mathbf{x} \mathbf{\hat{W}}_{\text{down}}.\\
    \end{aligned}
    \label{eq:fast}
\end{equation}

The resulting $\mathbf{C} \in \mathbb{R}^{n d_{\text{low}}}$ is then reshaped into an $n \times d_{\text{low}}$ matrix for subsequent computations.

In Section~\ref{sec:abl}, we demonstrate that an \name\ model achieves up to 97\% of the throughput of a traditional MoE model while also delivering superior downstream performance.

\begin{table*}[t]
    \caption{
    Ablations were performed on 732M-parameter language models (with 247M active parameters).
    Each model was trained on 100 billion tokens. 
    The results, highlighted in color, emphasize superior performance compared to configuration \circlednum{2}, the most common MoE setup. 
    Bold text indicates that the configuration outperforms the best traditional MoE variant in terms of average performance. 
    }
\vskip 0.1in
\begin{center}
\begin{small}
    \label{tab:abl}
    \begin{threeparttable}
    \begin{tabular}{l|cccccccc|c}
    \toprule
    Configuration & ARC-E & PIQA & SIQA & WINO & HELLA & MNLI & QNLI & SST2 & AVG. \\\midrule
     \circlednum{1} \  Traditional MoE                                  & 39.90 & 58.43 & 35.67 & \cellcolor{blue!7}{52.09} & 27.98 & \cellcolor{blue!7}{33.09} & 49.28 & 49.66 & 43.28  \\
     \circlednum{2} \quad + $\mathcal{L}_{\text{aux}}$      & 40.74 & 58.49 & 36.13 & 51.30 & 28.11 & 32.67 & 50.23 & 51.83 & 43.68 \\
    \circlednum{3} \quad + $\mathcal{L}_{\text{aux}}$ + Factorized $\mathbf{W}_{g}$      & 40.45 & \cellcolor{blue!7}{58.65} & \cellcolor{blue!7}{36.75} & \cellcolor{blue!7}{52.09} & 28.03 & 32.55 & 50.08 & 51.03  & 43.70  \\
    \circlednum{4} \quad + $\mathcal{L}_{\text{aux}}$ + Large Router                     & \cellcolor{blue!7}{41.41} & 57.62 & \cellcolor{blue!7}{36.64} & \cellcolor{blue!7}{52.33} & \cellcolor{blue!7}{28.34} & \cellcolor{blue!7}{33.18} & 49.53 & 50.69 & 43.71 \\\hline
    \circlednum{5} \ \name\ {\tiny($d_{\text{low}}=64$)}    & 39.77 & \cellcolor{blue!7}{58.71} & 35.31 & \cellcolor{blue!7}{52.33} & \cellcolor{blue!7}{28.29} & \cellcolor{blue!7}{32.78} & \cellcolor{blue!7}{50.27} & \cellcolor{blue!7}{52.98} & \textbf{43.81} \\
    \circlednum{6} \quad + $\mathcal{L}_{\text{aux}}$       & \cellcolor{blue!7}{42.17} & 57.67 & \cellcolor{blue!7}{36.75} & 50.75 & \cellcolor{blue!7}{28.15} & \cellcolor{blue!7}{34.06} & \cellcolor{blue!7}{50.49} & \cellcolor{blue!7}{53.10} & \textbf{44.12} \\\hline
    \circlednum{7} \ \name\ {\tiny($d_{\text{low}}=128$)}   & 40.70 & \cellcolor{blue!7}{59.41} & \cellcolor{blue!7}{36.64} & \cellcolor{blue!7}{52.09} & 28.06 & \cellcolor{blue!7}{34.38} & \cellcolor{blue!7}{50.69} & \cellcolor{blue!7}{53.21} & \textbf{44.39} \\
    \circlednum{8} \quad + $\mathcal{L}_{\text{aux}}$       & \cellcolor{blue!7}{41.33} & \cellcolor{blue!7}{58.65} & \cellcolor{blue!7}{36.80} & 50.75 & \cellcolor{blue!7}{28.40} & \cellcolor{blue!7}{33.71} & 49.55 & \cellcolor{blue!7}{53.10} & \textbf{44.04} \\\hline
    \circlednum{9} \ \name\ {\tiny($d_{\text{low}}=256$)}   & \cellcolor{blue!7}{41.08} & \cellcolor{blue!7}{58.81} & \cellcolor{blue!7}{36.44} & \cellcolor{blue!7}{51.70} & \cellcolor{blue!7}{28.23} & 32.24 & \cellcolor{blue!7}{50.54} & \cellcolor{blue!7}{53.90} & \textbf{44.12} \\
    \circledNum{10} \quad + $\mathcal{L}_{\text{aux}}$      & \cellcolor{blue!7}{41.16} & 58.32 & \cellcolor{blue!7}{36.80} & \cellcolor{blue!7}{53.04} & \cellcolor{blue!7}{28.37} & \cellcolor{blue!7}{32.78} & \cellcolor{blue!7}{50.61} & \cellcolor{blue!7}{54.59} & \textbf{44.46} \\\hline
    \circledNum{11} \name\ {\tiny($d_{\text{low}}=512$)} & 40.57 & 57.89 & \cellcolor{blue!7}{36.75} & 50.59 & \cellcolor{blue!7}{28.38} & \cellcolor{blue!7}{32.71} & 49.72 & \cellcolor{blue!7}{53.56} & \textbf{43.77} \\
    \circledNum{12} \quad + $\mathcal{L}_{\text{aux}}$      & \cellcolor{blue!7}{41.16} & 57.83 & \cellcolor{blue!7}{36.75} & \cellcolor{blue!7}{52.09} & \cellcolor{blue!7}{28.30} & \cellcolor{blue!7}{34.92} & \cellcolor{blue!7}{50.67} & 50.92 & \textbf{44.08} \\\bottomrule
    \end{tabular}
    \end{threeparttable}
    \end{small}
    \end{center}
\end{table*}

\section{Experiments}
\label{sec:exp}

We begin by providing a detailed analysis of our method through ablation experiments on pre-trained small language models using \name\ and traditional MoE. 
These experiments enable us to answer key research questions related to \name. 
Based on the insights gained, we scale up the language models to 4 billion parameters, demonstrating \name's scalability.

\subsection{Method Analysis through Small Language Models}
\label{sec:abl}

\subsubsection{General Setup}

We train small language models consisting of 12 layers, each containing 12 attention heads. 
Each layer contains 8 experts, with the top-$K=2$ experts selected. 
Models use the Llama~\cite{touvron2023llamaopenefficientfoundation} vocabulary of size 32,000 and the same pre-RMSNorm~\cite{zhang2019rootmeansquarelayer} module. 
We set \(d_{\text{model}} = 768\) and \(d_{\text{ffn}} = 3{,}072\) for traditional MoE models, while the values of \(d_{\text{low}}\) and \(d_{\text{wide}}\) for \name\ models are variable. 
Specifically, in all experiments below, to ensure that the total number of parameters in an \name\ model is comparable to that of an MoE model, when we adjust \(d_{\text{low}}\), \(d_{\text{wide}}\) is set as follows:
\begin{align}
d_{\text{wide}} = \frac{3 \cdot d_{\text{model}} \cdot d_{\text{ffn}} - d_{\text{low}} \cdot d_{\text{model}}}{d_{\text{low}} + 2 \cdot d_{\text{model}}}.
\label{eq:wide}
\end{align}
The total number of model parameters is 732 million, and the number of activated parameters is 247 million.

We train models on 100 billion tokens from RedPajama~\cite{together2023redpajama}, with a batch size of 4.2 million tokens, a learning rate of \(2 \times 10^{-4}\), and a linear warmup over the first 4,800 steps, followed by a cosine decay schedule that reduces the learning rate to \(1.28 \times 10^{-5}\)~\cite{StableLM-3B-4E1T}. 
The AdamW optimizer~\cite{adamw} is employed with \((\beta_1, \beta_2) = (0.9, 0.95)\), a gradient norm clipping threshold of 1, and a weight decay of 0.1.

We conduct a comprehensive evaluation of language models across a range of widely used tasks, including \textit{ARC-easy}~\cite{clark2018thinksolvedquestionanswering}, \textit{PIQA}~\cite{piqa}, \textit{SIQA}~\cite{sap-etal-2019-social}, \textit{Winogrande}~\cite{wino}, \textit{HellaSwag}~\cite{zellers-etal-2019-hellaswag}, \textit{MNLI}~\cite{mnli}, \textit{MRPC}~\cite{dolan-brockett-2005-automatically}, \textit{QNLI}~\cite{glue}, \textit{QQP}~\cite{glue}, and \textit{SST-2}~\cite{sst2}. 
The first five tasks are evaluated zero-shot, while the remaining tasks are tested three-shot because models exhibit unstable performance in zero-shot scenarios, with most errors arising from incorrect answer formats.
The accuracy is reported in Table \ref{tab:abl}.

\begin{figure}[t]
\begin{center}
\centerline{\includegraphics[width=\linewidth]{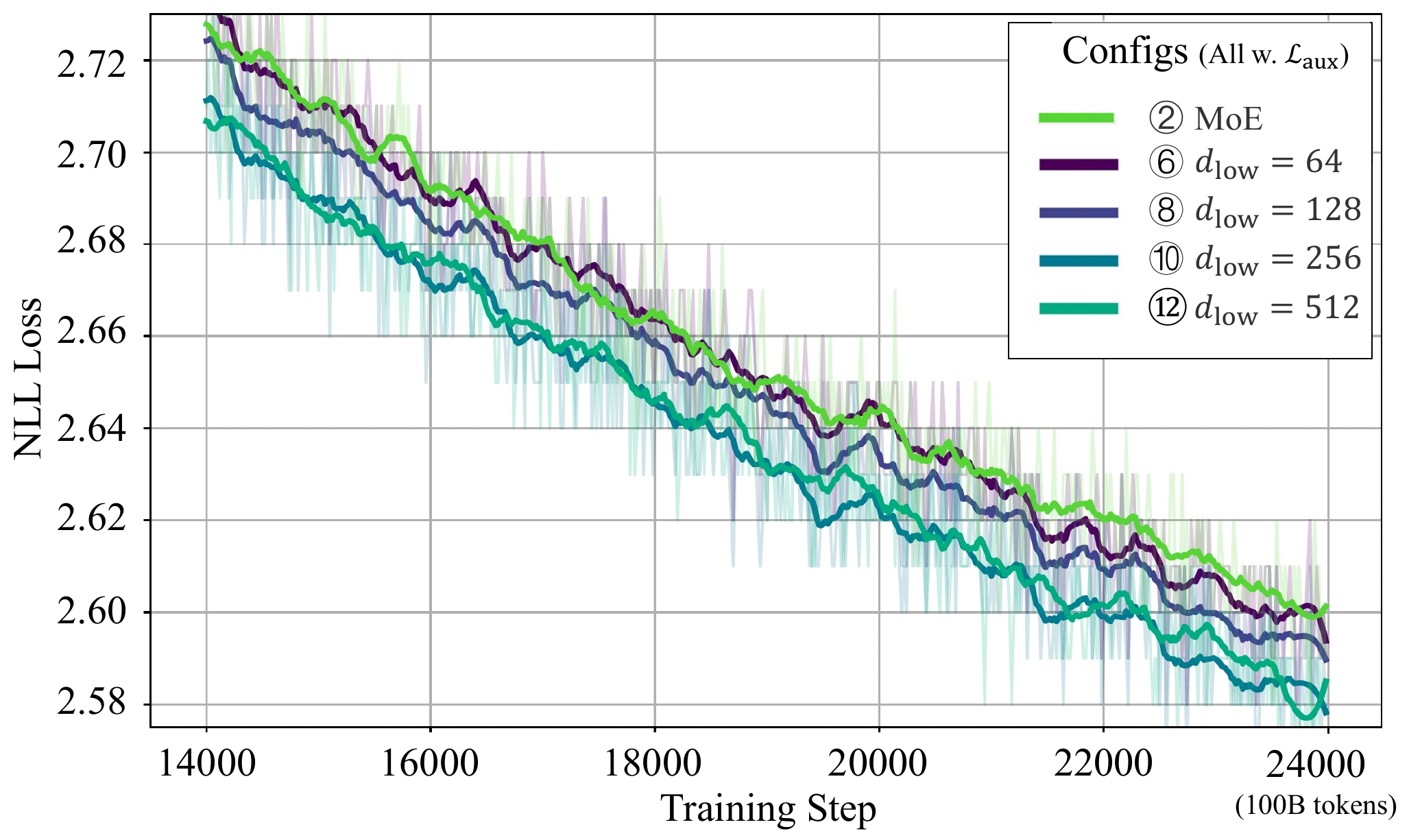}}
\caption{Pre-training NLL losses. 
All configurations shown are trained with $\mathcal{L}_{\text{aux}}$, though its value is not included in the figure.}
\label{fig:loss}
\end{center}
\end{figure}

\subsubsection{Resolving Questions Regarding \name}

We investigate the following questions related to \name\ through a series of ablation experiments.

\textbf{Question 1: How does the downstream performance of \name\ compare with traditional MoE models?} 
We evaluated various configurations of \name\ (Configs. \circlednum{5} to \circledNum{12}) and traditional MoE models (Configs. \circlednum{1} to \circlednum{4}). 
Every \name\ setup outperforms the best-performing MoE setup in terms of average accuracy across eight tasks. 
Notably, \name\ without any auxiliary loss surpasses traditional MoE models, which enhances the simplicity of training an MoE model.
Additionally, \name\ exhibits lower training loss, suggesting more efficient training. We elaborate on this in Question 2.

\begin{figure*}[t]
\begin{center}
\centerline{\includegraphics[width=\linewidth]{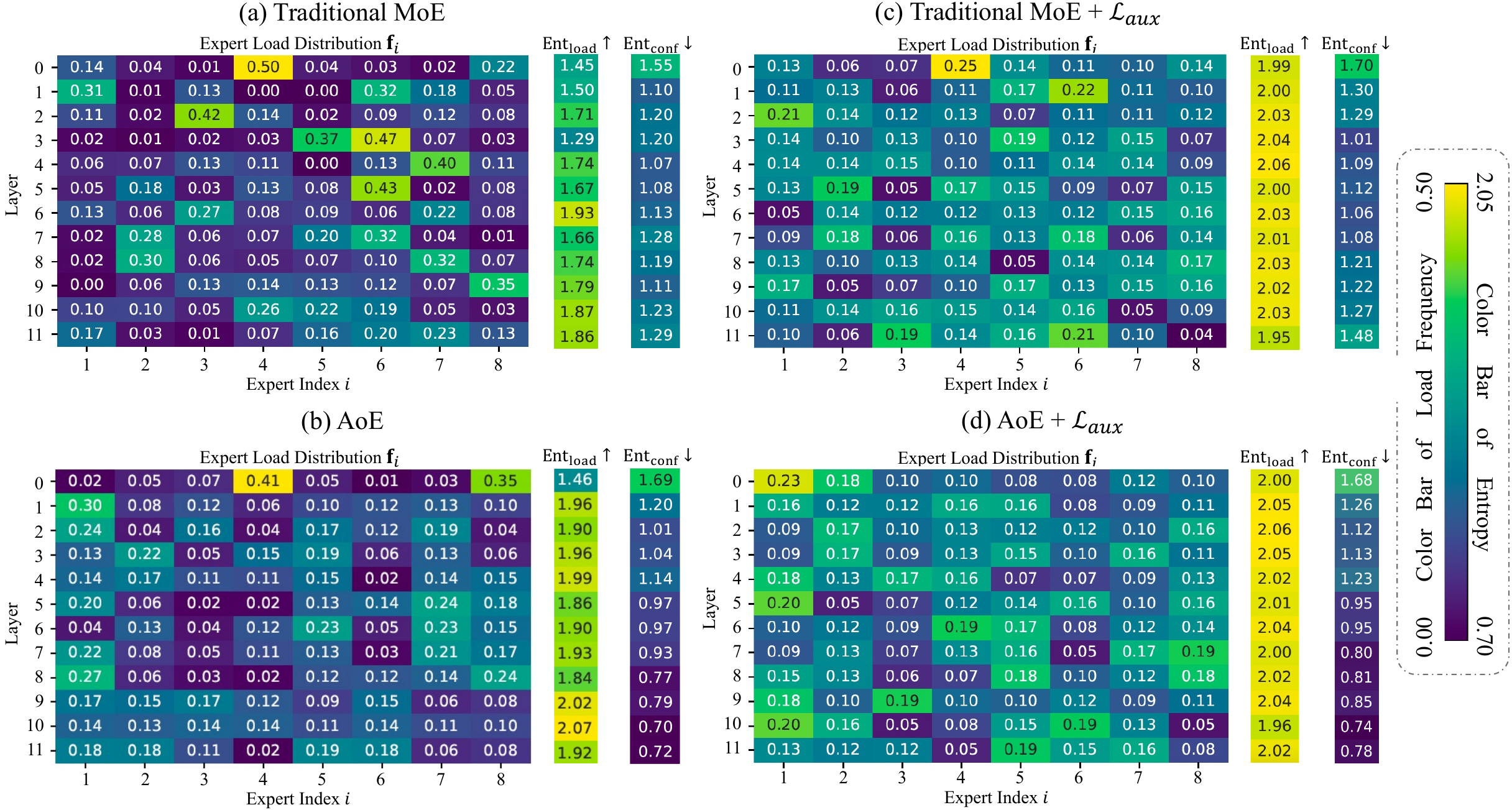}}
\caption{Statistical analysis of expert load.
The figure reveals several key insights: (1) $\mathcal{L}_{\text{aux}}$ enhances load balancing in both traditional MoE and \name .  
(2) \name s generally exhibit more balanced load distributions compared to their traditional MoE counterparts, as indicated by higher $\text{Ent}_{\text{load}}$ values. 
(3) \name s also demonstrate greater confidence in expert selection, reflected by lower $\text{Ent}_{\text{conf}}$ values.}
\label{fig:bl}
\end{center}
\end{figure*}

\textbf{Question 2: What is the impact of varying $d_{\text{low}}$?} 
We adjusted $d_{\text{low}}$ to values of 64, 128, 256, and 512, corresponding to Configs. \circlednum{6}, \circlednum{8}, \circledNum{10}, and \circledNum{12}, respectively.
The combined impact of $\mathcal{L}_{\text{aux}}$ and $d_{\text{low}}$ will be discussed in the next question.
All of these variants outperform the traditional MoE model in downstream performance. 
The performance differences among these configurations are relatively small. 
The maximum performance gain occurs when $d_{\text{low}}$ is approximately one-third of $d_{\text{model}}$ (256/768). 
Both smaller and larger values of $d_{\text{low}}$ result in lower performance, though they still surpass the baselines. 
The suboptimal performance with smaller $d_{\text{low}}$ may be due to the factorization of $\textbf{W}_{g}$ into $\textbf{W}_{\text{down}}\textbf{W}_{\text{up}}$ being a lossy approximation when $d_{\text{low}}$ is below the true rank of $\textbf{W}_{g}$. 
Conversely, larger $d_{\text{low}}$ introduce more noise into the activation, potentially hindering the effectiveness of the norm-based selection measure.

In Figure~\ref{fig:loss}, we present the negative log-likelihood (NLL) loss during training for traditional MoE (Config. \circlednum{2}) and AoE models (Configs. \circlednum{6}, \circlednum{8}, \circledNum{10}, and \circledNum{12}). 
AoE models exhibit more effective expert learning, as evidenced by lower loss values. 
However, when $d_{\text{low}} = 64$ (Config. \circlednum{6}), the loss is comparable to that of traditional MoE models, suggesting that smaller $d_{\text{low}}$ values hinder AoE performance. 
In contrast, $d_{\text{low}} = 256$ (Config. \circledNum{10}) results in the lowest training loss overall, reinforcing the finding that setting $d_{\text{low}}$ to approximately one-third of $d_{\text{model}}$ yields the most benefits.

\textbf{Question 3: How is the load balancing of \name?}

There are three main findings regarding load balancing.

\textbf{\textit{Finding 3.1:}} \textit{\name\ improves load balancing compared to traditional MoE models, with or without $\mathcal{L}_{\text{aux}}$.}

\name\ can incorporate $\mathcal{L}_{\text{aux}}$ with minor modifications to Eq.~\ref{eq:aux}, as shown below:
\begin{equation}
    \begin{aligned}
        &\mathcal{L}_{\text{aux}} =\alpha_{\text{aux}} \cdot n \cdot \sum_{i=1}^{n} \mathbf{f}_{i} \cdot \mathbf{P}_{i}, \text{ where} \\
    \end{aligned}
    \label{eq:aux2}
\end{equation}
\begin{align*}
    \mathbf{f}_{i} = \frac{1}{T} \sum_{\mathbf{x} \in \mathcal{B}} \mathbbm{1} \left\{ i \in \argtopk \left( \texttt{L2-Norm} \left( \mathbf{xW}^{i}_{\text{down}} \right) \right) \right\},
\end{align*}
\begin{align*}
    \mathbf{P}_{i} = \frac{1}{T} \sum_{\mathbf{x} \in \mathcal{B}} \texttt{Softmax} \left( \texttt{L2-Norm} \left( \mathbf{xW}^{i}_{\text{down}} \right) \right)[i],
\end{align*}
and $\alpha_{\text{aux}}$ is determined using a validation set comprising 5 billion tokens from~\cite{Gokaslan2019OpenWeb}. 
Experiments indicate that $\alpha_{\text{aux}} = 0.01$ is effective for both traditional MoE and \name\ models. 
We adopted this value across all configurations without further hyperparameter tuning. 

Figure~\ref{fig:bl} illustrates expert load statistics on the SST-2 dataset~\cite{sst2} for Configs. \circlednum{1}, \circlednum{2} (Traditional MoE with and without $\mathcal{L}_{\text{aux}}$), \circlednum{9}, and \circledNum{10} (\name\ with and without $\mathcal{L}_{\text{aux}}$). 
We report both the load distribution $\mathbf{f}_{i}$ (as defined in Eqs.~\ref{eq:aux} and\ref{eq:aux2}), representing the percentage of tokens processed by expert $i$, and the entropy of the load distribution within each layer: 
\begin{equation}
\begin{aligned}
\text{Ent}_{\text{load}} = -\sum^{n}_{i=1} \mathbf{f}_{i} \log \mathbf{f}_{i}. 
\end{aligned}
\end{equation}
Higher entropy values indicate more balanced load distributions across experts.
Comparing Figures~\ref{fig:bl}(a) and~\ref{fig:bl}(b), without $\mathcal{L}_{\text{aux}}$, \name\ achieves a more balanced load distribution in 11 out of 12 layers.
Comparing Figures~\ref{fig:bl}(c) and~\ref{fig:bl}(d), with $\mathcal{L}_{\text{aux}}$, \name\ maintains a superior overall balance.
For reference, the average $\text{Ent}_{\text{load}}$ values for subfigures (c) and (d) are 2.015 and 2.023, respectively.
\footnote{$d_{\text{low}}$ has minimal impact on the statistical metrics discussed in Question 4. As a result, we do not provide analysis for other configurations, as they offer little additional insight.}

\textbf{\textit{Finding 3.2:}} \textit{\name\ models exhibit stronger confidence in expert selection.}

We introduce the confidence entropy, denoted as $\text{Ent}_{\text{conf}}$.
For each layer, we have:
{\begin{equation}
\begin{aligned}
    &\text{Ent}_{\text{conf}} = -\sum_{i=1}^{n} \mathbf{p}_{i} \log \mathbf{p}_{i},\\
    &\mathbf{p}_{i} = \left\{
    \begin{aligned}
    &\texttt{Softmax}\left(\texttt{L2-Norm}\left(\mathbf{x} \mathbf{W}^{i}_{\text{down}}\right)\right) \text{, for \name}\\
    &\texttt{Softmax}\left(R(\mathbf{x})\right) \text{, for traditional MoE}
    \end{aligned}
    \right. 
\end{aligned}
\label{eq:conf}
\end{equation}}

This entropy quantifies the confidence in expert selection: lower entropy indicates a distribution closer to a one-hot vector, signifying more confident expert selection, while higher entropy reflects greater uncertainty in expert decisions.
\name\ exhibits significantly lower entropy, demonstrating stronger confidence in selecting experts. 
Furthermore, its confidence increases from shallow to deep layers, aligning with the intuitive inductive bias that shallow layers perform fundamental, non-specialized functions, whereas deeper layers handle specialized and abstract tasks~\cite{wang2023interpretability,lv2024interpretingkeymechanismsfactual}.
In contrast, MoE models do not display this trend, potentially suggesting more homogeneous expertise within and across layers~\cite{wang2024hmoeheterogeneousmixtureexperts}.

\textbf{\textit{Finding 3.3:}} \textit{Beyond improved load balancing, \name\ with $\mathcal{L}_{\text{aux}}$ achieves better downstream performance.}

In general, $\mathcal{L}_{\text{aux}}$ benefits both traditional MoE and \name\ models. 
However, when $d_{\text{low}} = 128$, applying $\mathcal{L}_{\text{aux}}$ results in a decrease in accuracy, which we attribute to task-specific variations.
In conclusion, as addressed in response to Question 4, \name\ exhibits strong potential for advancing MoE-based LLMs, owing to its improvements in both load balancing and downstream performance.

\begin{table*}[t]
\caption{
Comparison of traditional MoE and AoE models trained using alternative expert-selection strategies. 
For the Top‑$P$ strategy, the number of activated parameters is input-dependent but nearly the same between the two models, whereas the expert-choice strategy activates 247 out of 732M parameters.
}
\label{tab:strategy}
\vskip 0.1in
\begin{center}
\begin{small}
\begin{tabular}{c|c|cccccccc|c}
    \toprule
   Strategy & Model & ARC-E & PIQA & SIQA & WINO & HELLA & MNLI & QNLI & SST2 & AVG. \\\midrule
   Top-$P$ & Traditional MoE  & 41.08 & 57.96 & 37.46 & 50.36	& 28.25 &			32.79 & 50.39 & 52.64 & 43.87 \\
   \cite{huang2024hardertasksneedexperts} & \name & 41.04 &	\cellcolor{blue!7}{58.65} &	36.39 &	\cellcolor{blue!7}{51.07} &	\cellcolor{blue!7}{28.35} & \cellcolor{blue!7}{32.96} &	\cellcolor{blue!7}{51.46} &	\cellcolor{blue!7}{54.36} & \textbf{44.29} \\\midrule
    Expert-Choice & Traditional MoE  & 40.91 & 59.09 & 37.26 & 50.75 & 28.09 & 32.11 & 50.12 & 52.75 & 43.89 \\
   \cite{zhou2022mixtureofexperts} & \name  & \cellcolor{blue!7}{41.58} & 58.22 & 37.21 & \cellcolor{blue!7}{53.04} & \cellcolor{blue!7}{28.44} & \cellcolor{blue!7}{33.83} & \cellcolor{blue!7}{50.54} & 50.46 & \textbf{44.17}\\\bottomrule
\end{tabular}
\end{small}
\end{center}
\end{table*}

\begin{figure}[t]
\begin{center}
\centerline{\includegraphics[width=\linewidth]{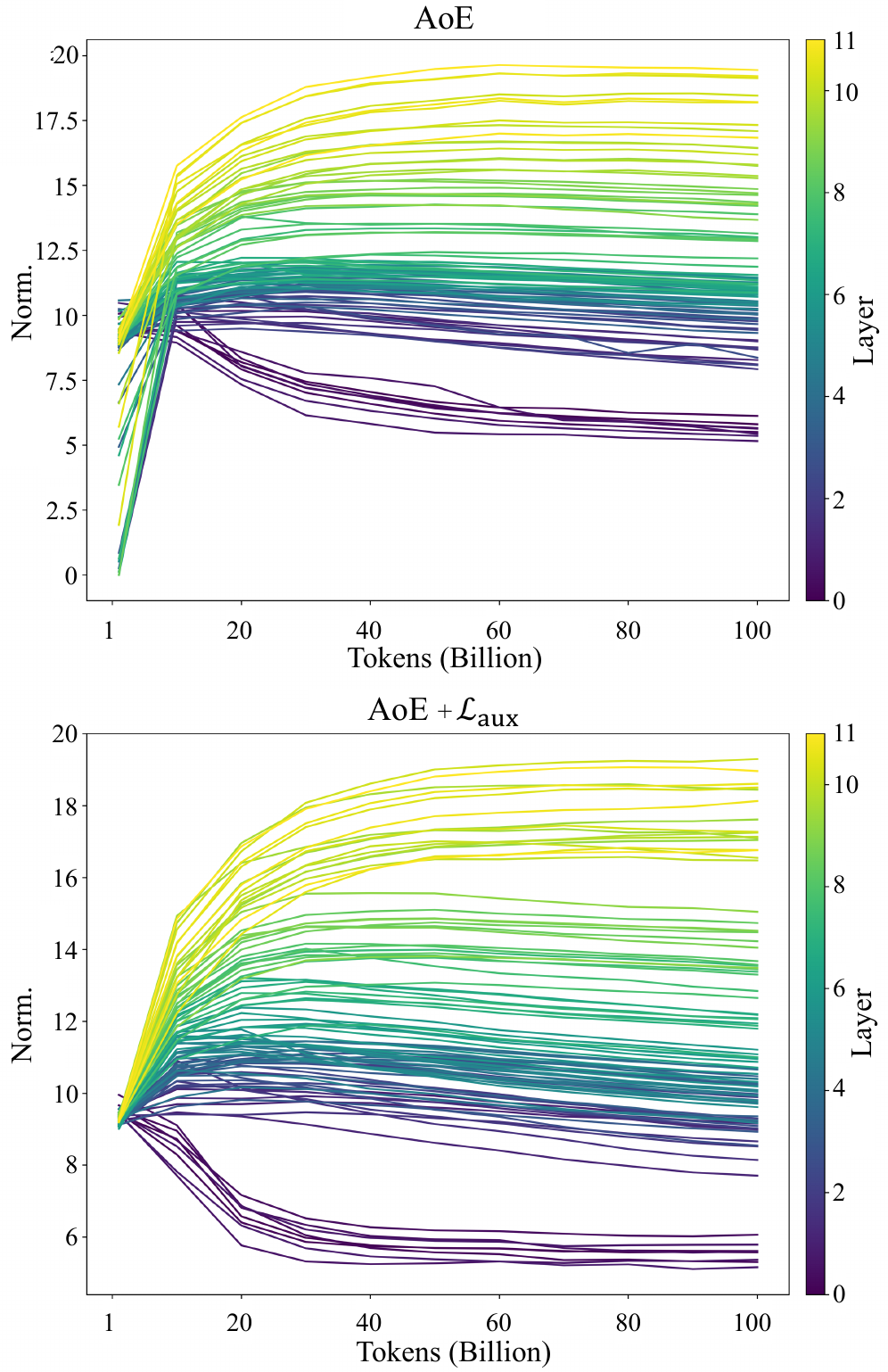}}
\caption{Average activation norm dynamics during training.
Each plot represents an expert, distinguished by color according to its layer.
Experts within the same layer achieve similar activation scales, indicating that their self-evaluation criteria for determining whether they are capable of processing inputs are aligned.}
\label{fig:cluster}
\end{center}
\end{figure}

\textbf{Question 4: Do improvements stem from the factorization of $\mathbf{W}_{g}$?} 
We examined the impact of factorizing the experts' weight matrix on performance by comparing Configurations \circlednum{3} and \circlednum{2}. 
The factorization does not significantly influence performance, as expected in Section~\ref{sec:method}, based on findings that the weights of LLMs are inherently low-rank~\cite{li2018measuring, aghajanyan-etal-2021-intrinsic, hu2022lora}. 
Therefore, the improvements observed with \name\ are not attributed to the factorization of model weights.

\textbf{Question 5: Does the improvement of \name\ come from involving more parameters in expert selection?} 
We increased the size of the router in MoE to include $n \cdot d_{\text{low}} \cdot d_{\text{model}}$ parameters, ensuring that the number of parameters involved in expert selection remains consistent with that of \name\ models. 
Note that in this setup, traditional MoE models have more activated parameters in total.
Comparing Config. \circlednum{4} and \circlednum{2}, the larger router provides a slight performance benefit.
However, every \name\ setup still outperforms this configuration. 
Thus, the improvement in \name\ is not primarily due to involving more parameters in expert selection.

\textbf{Question 6: How aligned are the self-evaluation criteria among experts?} 
In \name\ models, each expert independently develops self-evaluation criteria for processing tokens, as reflected in their activation scales.
This might raise concerns that some experts could become overly ``egoistic,'' meaning their internal activations are consistently larger than those of others. 
For example, one expert might produce activations with norms ranging from 10 to 20, while an ``ego'' expert produces activations with norms from 20 to 30, leading to biased selections that favor the ``ego'' expert.

We track dynamics of activation norms during pre-training.
Figure~\ref{fig:cluster} shows the details for Configs. \circlednum{9} and \circledNum{10}.
Except for the very initial period, experts' self-evaluation criteria are well aligned, as evidenced by clusters of same-colored plots (representing experts within the same layer).
In the early stages of training without the auxiliary loss, some middle-to-upper-layer experts exhibit significantly lower activation. 
However, \name\ naturally resolves this imbalance in activation scales during training. 
Alternatively, $\mathcal{L}_{\text{aux}}$ can address this imbalance earlier because it acts as a regularizer for activation norms, increasing the norm scales of underactive experts and ensuring they are used more often.

\textbf{Question 7: Is \name\ compatible with other expert-selection strategies?}
We also train language models using the Top-$P$ token-choice~\cite{huang2024hardertasksneedexperts} and the Top-$K$ expert-choice strategy~\cite{zhou2022mixtureofexperts}.

For Top-$P$ token-choice~\cite{huang2024hardertasksneedexperts}, we replace the Top-$K=2$ strategy with Top-$P=0.6$ following~\cite{wang2024hmoeheterogeneousmixtureexperts}. 
Models utilizing the Top-$P$ strategy require an additional auxiliary loss equivalent to minimizing our introduced $\text{Ent}_{\text{conf}}$ (Eq.~\ref{eq:conf}). 
This ensures that the model does not learn shortcuts by assigning uniform probabilities to all experts, which would activate too many parameters to achieve lower loss. 
Following~\cite{huang2024hardertasksneedexperts}, we set the weight of this regularization term to $10^{-4}$.

Expert-choice~\cite{zhou2022mixtureofexperts} is similar to the Top-$K$ token-choice strategy. 
Consider an expert-selection matrix in the shape of $T \times n$ (i.e., the router outputs in traditional MoE or the activation norms in \name). 
The token-choice strategy applies the Top-$K$ operator along the $n$ dimension, whereas expert-choice applies it along the $T$ dimension.
Models trained using the expert-choice strategy do not require auxiliary losses. 
We set the ``capacity factor'' to 2 (see~\cite{zhou2022mixtureofexperts} for details), allowing each expert to process 25\% of the tokens in a batch.

Results are shown in Table~\ref{tab:strategy}, where $d_{\text{low}}$ in these experiments is 256.
\name\ outperforms traditional MoE models, demonstrating its generality across various expert-selection strategies.

\begin{table}[t]
\caption{Throughput and memory usage comparison among several configurations. 
Auxiliary losses do not impact efficiency.}
\label{tab:eff}
\vskip 0.1in
\begin{center}
\begin{small}
\begin{tabular}{l|c}
    \toprule
    Configuration & TP. (K/s) / Mem. (GB) \\\midrule
     Traditional MoE & 51.42 / 50.61\\
     \name\ {\tiny($d_{\text{low}}=64$)} & 49.79 / 59.39\\
     \name\ {\tiny($d_{\text{low}}=128$)} & 49.42 / 57.86 \\
     \name\ {\tiny($d_{\text{low}}=256$)} & 47.98 / 57.32 \\
     \name\ {\tiny($d_{\text{low}}=512$)} & 46.07 / 55.90 \\\bottomrule
\end{tabular}
\end{small}
\end{center}
\end{table}

\begin{table*}[t]
\caption{For 4B-parameter LLMs (with 1.18B active parameters), \name\ exhibits better downstream performance than MoE models.}
\label{tab:llm}
\vskip 0.1in
\begin{center}
\begin{small}
\begin{tabular}{c|cccccccc|c}
    \toprule
    Model & ARC-E & PIQA & SIQA & WINO & HELLA & MNLI & QNLI & SST2 & AVG. \\\midrule
     Traditional MoE & 53.70 & 65.40 & 39.10 & 51.54 & 35.80 & 32.19 & 49.77 & 57.00 & 48.06 \\
     \name  & \cellcolor{blue!7}{55.98} & \cellcolor{blue!7}{65.61} & \cellcolor{blue!7}{39.87} & \cellcolor{blue!7}{52.57} & \cellcolor{blue!7}{36.77} & \cellcolor{blue!7}{35.39} & \cellcolor{blue!7}{50.05} & \cellcolor{blue!7}{61.93} & \textbf{49.80} \\\bottomrule
\end{tabular}
\end{small}
\end{center}
\end{table*}

\textbf{Question 8: How Efficient is \name?\ }
Table~\ref{tab:eff} shows the maximum training throughput (tokens processed per second per GPU) and memory usage for both traditional MoE models and various \name\ models. 
Here are the key findings:

\textbf{\textit{Finding 8.1:}} \textit{\name\ achieves up to 97\% of the throughput of the traditional MoE model, with the added cost of memory.}

Additionally, note that experts in our experiments work sequentially within the same layer but in practical deployments of MoE-LLMs, experts are typically distributed across different devices and operate in parallel. 
Consequently, experts must wait for the most loaded expert to finish computation, resulting in idle time that can be quantified by the difference between the maximum and minimum expert loads. 
The total differences across layers are 1.49 for Figure~\ref{fig:bl}(c) (traditional MoE) and 1.41 for Figure~\ref{fig:bl}(d) (\name).
In this case, \name\ can achieve an additional time reduction equivalent to processing 8\% of the total tokens through a single MoE layer.
Assuming an ideal load distribution where each of the 8 experts processes 12.5\% of the total tokens, this reduction translates to a 64\% decrease in the running time of one MoE layer.
This advantage, however, is not reflected in the reported efficiency metrics.

\textbf{\textit{Finding 8.2:}} \textit{In \name, memory usage and throughput are influenced by \( d_{\text{low}} \), presenting trade-offs.}

In terms of incremental memory, a smaller \( d_{\text{low}} \) requires a larger \( d_{\text{wide}} \), thereby increasing the memory consumption of \( \mathbf{xW}_{\text{up}} \) to \( T \cdot d_{\text{wide}} \), where \( T \) is the number of tokens.
Conversely, a larger \( d_{\text{low}} \) results in a larger activation cache, raising memory usage to \( n \cdot T \cdot d_{\text{low}} \).
For Configs.~\circlednum{6} to \circledNum{10}, \( n \cdot d_{\text{low}} < d_{\text{wide}} \), making the primary memory cost stem from the larger up-projection. 
In contrast, Config.~\circledNum{11} and \circledNum{12} satisfy \( n \cdot d_{\text{low}} > d_{\text{wide}} \), meaning the increased memory usage is more attributable to the larger activation cache.
In terms of throughput reduction, a smaller \( d_{\text{low}} \) requires more computational resources for the up-projection, while a larger \( d_{\text{low}} \) leads to a higher unused activation cache.
It is worth noting that the efficiency of AoE diminishes as the number of experts increases and as sparsity grows. 
We are actively working on further optimizing AoE's efficiency under these conditions.

\subsection{Pre-training Large Language Models}
\label{sec:llm}

We pre-train LLMs with a total of 4 billion parameters, of which 1.18B are activated. 
The initial learning rate is $3.2 \times 10^{-4}$~\cite{StableLM-3B-4E1T}. 
Each model has 24 layers, with 20 attention heads per layer. 
For traditional MoE models, we set \(d_{\text{model}} = 1{,}280\) and \(d_{\text{ffn}} = 5{,}120\).
Considering the trade-offs between efficiency overhead and performance gain, we set \(d_{\text{low}} = 400\) and, according to Eq.~\ref{eq:wide}, derive \(d_{\text{wide}} = 6{,}470\). 
Other settings follow those in Section~\ref{sec:abl}.
Both models are enhanced by $\mathcal{L}_{\text{aux}}$ with $\alpha_{\text{aux}} = 0.01$. 
Table~\ref{tab:llm} demonstrates that \name\ outperforms traditional MoE models as they scale, with the performance improvement being more pronounced in LLMs compared to smaller models.
This highlights the potential of \name\ to drive advancements in larger and more powerful MoE-based LLMs.

\section{Conclusion}
We introduce \fullname\ (\name), a novel Mixture-of-Experts (MoE) paradigm that addresses a crucial yet widely overlooked issue: the separation between the router's decision-making and the experts' execution, which leads to suboptimal expert selection and learning. 
\name\ selects experts based on their internal activation scales. 
Several architectural modifications ensure efficiency. 
Language models based on \name\ outperform traditional MoE models in many aspects. 
This paper highlights the advantages of enabling MoE experts to self-select and aims to inspire the community to develop more powerful MoE-like models.

\section*{Acknowledgement}

Ang Lv is supported by the Outstanding Innovative Talents Cultivation Funded Programs 2023 of Renmin University of China and CIE-Tencent Doctoral Student Research Incentive Program (HunYuan Large Language Model Special Project).
Ruobing Xie is supported by the Young Elite Scientists Sponsorship Program by CAST (2023QNRC001).
This work is also supported by the Public Computing Cloud, Renmin University of China and by fund for building world-class universities (disciplines) of Renmin University of China.

\section*{Impact Statement}
Training large language models can generate content with ethical implications. 
Many effective techniques can align the preferences and values of LLMs to mitigate these concerns. 
Beyond this, we believe that our work does not introduce additional societal or ethical issues.

\bibliography{main}
\bibliographystyle{icml2025}


\newpage
\appendix
\onecolumn
\section{Re-running Experiments in Section~\ref{sec:ob} Using Alternative Expert-Selection Metrics}
\label{apx:norms}

We also use the $L^{1}$ and $L^{\infty}$ norms as expert-selection metrics in pre-trained LLMs, which resulted in poorer performance preservation compared to the $L^{2}$ norm. 
The time costs for each configuration are identical to those presented in Table~\ref{tab:pre} and are therefore omitted here for clarity. 
The results are shown below.

\begin{table*}[h]
    \caption{Preliminary study results on pre-trained MoE-LLMs, selecting experts by $L^{1}$ norm of internal activation.}
    \label{tab:pre-l1}
    \vskip 0.1in
\begin{center}
\begin{small}
    \begin{tabular}{c|c|c|c|c}
    \toprule
    \multirow{2}{*}{\shortstack{Node for Norm\\ Calculation}} & \multicolumn{2}{c|}{MMLU (5-shot)} & \multicolumn{2}{c}{ARC-C (5-shot)} \\\cmidrule(lr){2-5}
    & Mixtral $8\times7$B  & Phi-3.5-MoE-ins. & Mixtral $8\times7$B & Phi-3.5-MoE-ins. \\\hline
    $\mathbf{x}\mathbf{W}_{g}$  & 51.14 & 24.15  & 41.98  & 29.01  \\
    $\mathbf{x}\mathbf{W}_{p}$ & 39.79 & 35.87 & 40.19 & 36.35 \\
    $\texttt{SiLU}(\mathbf{x}\mathbf{W}_{g})$ & 47.29 & 26.37  & 45.73  & 36.09  \\
    $\texttt{SiLU}(\mathbf{x}\mathbf{W}_{g}) \odot \mathbf{x}\mathbf{W}_{p}$ & 54.37 & 26.95 &  50.09 & 33.79  \\
    Experts' Final Outputs & 57.84 & 26.56 & 52.73 & 31.31  \\\midrule
    \midrule
    Performance w. Router & 70.35  & 78.20 & 62.12 & 67.41 \\\bottomrule
    \end{tabular}
    \end{small}
    \end{center}
\end{table*}

\begin{table*}[h]
    \caption{Preliminary study results on pre-trained MoE-LLMs, selecting experts by $L^{\infty}$ norm of internal activation.}
    \label{tab:pre-linf}
    \vskip 0.1in
\begin{center}
\begin{small}
    \begin{tabular}{c|c|c|c|c}
    \toprule
    \multirow{2}{*}{\shortstack{Node for Norm\\ Calculation}} & \multicolumn{2}{c|}{MMLU (5-shot)} & \multicolumn{2}{c}{ARC-C (5-shot)} \\\cmidrule(lr){2-5}
    & Mixtral $8\times7$B  & Phi-3.5-MoE-ins. & Mixtral $8\times7$B & Phi-3.5-MoE-ins.\\\hline
    $\mathbf{x}\mathbf{W}_{g}$  & 48.16 & 29.28 & 43.77  & 35.92  \\
    $\mathbf{x}\mathbf{W}_{p}$ & 50.43 & 34.78 & 49.49 & 40.02 \\
    $\texttt{SiLU}(\mathbf{x}\mathbf{W}_{g})$ & 54.30 & 36.38  & 47.95  & 50.85  \\
    $\texttt{SiLU}(\mathbf{x}\mathbf{W}_{g}) \odot \mathbf{x}\mathbf{W}_{p}$ & 50.72 & 26.43 & 46.08  & 33.02  \\
    Experts' Final Outputs & 51.03 & 23.64  & 53.16 & 30.12  \\\midrule
    \midrule
    Performance w. Router & 70.35  & 78.20 & 62.12 & 67.41 \\\bottomrule
    \end{tabular}
    \end{small}
    \end{center}
\end{table*}

\section{Additional Interpretation of \name's Advantage}
\label{apx:toy}

We provide some intuitive insights into \name's strengths by developing a fully controlled classification task and monitoring training dynamics of both tiny \name\ and MoE models.
We provide details here for interested readers.
This experiment is of a toy nature and not intended as a major claim or contribution.

\begin{figure*}[t]
\begin{center}
\centerline{\includegraphics[width=\linewidth]{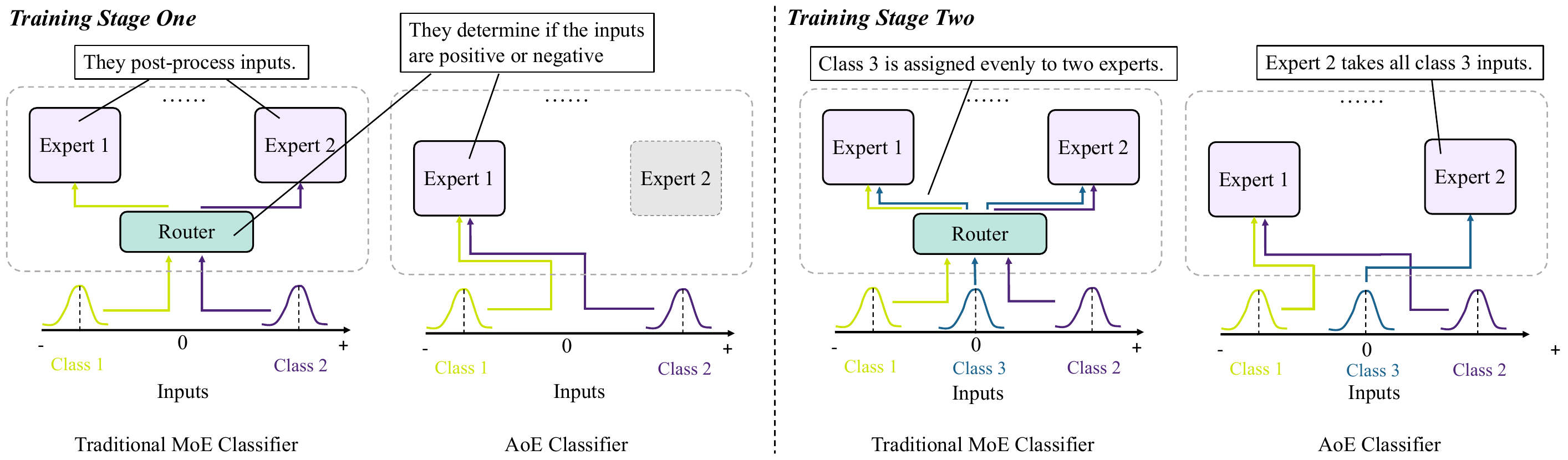}}
\caption{The overview of our toy experiments training tiny \name\ and traditional MoE classifiers.}
\label{fig:toy}
\end{center}
\end{figure*}

In our setup, inputs are multivariate Gaussian vectors belonging to three classes. 
Classes one and two have distinct positive and negative means, respectively, while class three has a zero mean. 
We adjust their standard deviations to ensure no overlap within a three-sigma range. Initially, we train both tiny \name\ and MoE classifiers to distinguish between classes one and two; this is referred to as training stage one.
After convergence, we introduce class three into the training process and continue training, referred to as training stage two. 
The classifiers consist of a single layer with two experts. 
Throughout training, we monitor expert behaviors, such as internal activation scales and token load.
Figure~\ref{fig:toy} illustrates the pipeline and results of this toy experiment.

During training stage one, we observed that MoE classifiers assign class one and class two to different experts. 
This suggests that the classification role is primarily handled by the router, while the experts perform post-processing. 
In contrast, \name\ uses only one expert to process all inputs during training stage one. 
Early in training, one expert identifies that the two classes are separable and develops the capability for binary classification. 
As training progresses, this expert's ability (reflected in increasingly larger activation norms) causes the other expert to remain naturally idle.

In training stage two, MoE evenly assigns inputs from the newly added class three to both experts.
This occurs because the router has been trained for binary classification and lacks the capacity to handle out-of-distribution inputs, leading to equal prediction distribution across experts. 
This exacerbates the issue of homogeneous experts in the MoE classifier, as the capability to classify class three is also distributed across all experts.
Conversely, in the \name\ classifier, the expert handling classes one and two exhibits low activation when presented with third-class inputs.
Its activation is even lower than that of the idle expert, which lacks specialization and does not resist class three inputs. 
As a result, the idle expert naturally handles all class three inputs.
This results in heterogeneous experts within the \name\ classifier: one expert manages the negative-positive classification, while the other processes zero-mean inputs.

Notably, in these toy experiments, the expert load during the first training stage is not balanced in \name. In contrast, real-world pre-trained language models do not exhibit this imbalance, as shown in Figure~\ref{fig:bl}.
The reason is that the classification of input features in practical scenarios is far more complex, with a greater number of classes involved.
As an evidence, when class three is added during training, \name\ achieves a balanced expert load.

Comparing token assignments between the two models reveals several drawbacks of traditional MoE models:

(1) Sub-optimal expert selection: 
The binary classification task of distinguishing between classes one and two, which is relatively easy, could be effectively managed by a single MLP (i.e., one expert). 
However, MoE classifiers utilize both experts due to the router's classification behavior.
This leads to under-exploitation of parameters and highlights the sub-optimal selection of experts in traditional MoE models, resulting from ``the separation between the router's decision and the experts' execution.''

(2) Distributed expertise: The ability to perform binary classification is distributed across two experts, preventing specialization.

The observation holds and near-zero loss is achieved as long as there is no overlap within a three-sigma range. 
In our experiments, we tested input dimensions and the model's $d_{\text{ffn}}$ and $d_{\text{wide}}$ parameters within the range of 32 to 256. 
When the input dimension is too small relative to the model dimension, the task becomes too easy to learn, and the above behavior is not observed. 
Conversely, if the input dimension is too large, the task becomes too difficult, preventing the loss from decreasing and rendering observed behavior uninformative.

\end{document}